\documentclass[runningheads]{llncs}
\usepackage{graphicx}
\usepackage{amsmath}
\usepackage{enumerate}
\usepackage{booktabs}
\usepackage{multirow}
\usepackage{hhline}
\usepackage{amssymb}
\usepackage{color}
\DeclareMathOperator*{\argmin}{arg\,min}

\usepackage[skip=5pt]{caption} % decrease space between figure and caption
\usepackage[misc,geometry]{ifsym} % \Letter symbol, a small envelope after corresponding author

\usepackage{lipsum}
\newcommand\blfootnote[1]{%
  \begingroup
  \renewcommand\thefootnote{}\footnote{#1}%
  \addtocounter{footnote}{-1}%
  \endgroup
}
\definecolor{darkgreen}{rgb}{0,0.6,0.3}

\begin{document}
\title{DeepCenterline: a Multi-task Fully Convolutional Network for Centerline Extraction}
\titlerunning{DeepCenterline for Centerline Extraction}

%\author{Zhihui Guo\inst{1}\orcidID{0000-1111-2222-3333} \and
%Second Author\inst{2,3}\orcidID{1111-2222-3333-4444} \and
%Third Author\inst{3}\orcidID{2222--3333-4444-5555}}
\author{Zhihui Guo*\inst{1} \and
Junjie Bai*\inst{2}$^{\text{(\Letter)}}$ \and
Yi Lu\inst{2} \and
Xin Wang \inst{2} \and
Kunlin Cao \inst{2} \and
Qi Song \inst{2} \and
Milan Sonka \inst{1} \and 
Youbing Yin \inst{2} 
}
\authorrunning{Z. Guo, J. Bai et al.}

\institute{
University of Iowa, Iowa City IA 52242, USA \and
CuraCloud Corporation, Seattle WA 98104, USA \\
\email{junjieb@curacloudcorp.com} 
}

%\institute{Princeton University, Princeton NJ 08544, USA \and
%Springer Heidelberg, Tiergartenstr. 17, 69121 Heidelberg, Germany

%
\maketitle              % typeset the header of the contribution
\begin{abstract}
%The abstract should briefly summarize the contents of the paper in
%\zzz{reworded -  still correct?}
A novel centerline extraction framework is reported which combines an end-to-end trainable multi-task fully convolutional network (FCN)  with a minimal path extractor. The FCN  simultaneously computes centerline distance maps and detects branch endpoints. The method generates single-pixel-wide centerlines with no spurious branches. It handles arbitrary tree-structured object with no prior assumption regarding depth of the tree or its bifurcation pattern. It is also robust to substantial scale changes across different parts of the target object and minor imperfections of the object's segmentation mask. To the best of  our knowledge, this is the first deep-learning based centerline extraction method that guarantees single-pixel-wide centerline for a complex tree-structured object.

%The proposed method generate a single-pixel-wide centerline for a tree-structure object with no prior assumptions regarding the tree structure such as the depth of tree or bifurcation pattern. The generated centerline is also free of incorrect spurious branches commonly generate by the traditional morphology based skeletonization algorithm. To our best knowledge, this is the first deep-learning based centerline extraction method. Our method handles arbitrary tree-structured object, and is robust to scale changes of object and minor imperfections of segmentation mask. 
%We proposed a novel end-end trainable multi-task fully convolutional network (FCN) with attention mechanism to
%While minimal cost path based centerline extraction methods have achieved quite some success in medical imaging applications, their two major inputs\textendash{--}cost metrics and end point list\textendash{--}still suffer from scale changes of target object and imperfections of segmentation mask, and require human intervention more or less. 
%To address these limitations, this paper presents a novel end-to-end trainable multi-task fully convolutional network (FCN) with attention mechanism to simultaneously compute centerline distance maps, and a complete list of end points of the target object skeleton. To our best knowledge, this is the first deep-learning based method for centerline extraction which generates tree-structured centerline without any assumption on number of branches or bifurcation pattern. 

The proposed method is validated in coronary artery centerline extraction on a dataset of 620 patients (400 of which used as test set). This application is challenging due to the large number of coronary branches, branch tortuosity,  and large variations in length, thickness, shape, etc. The proposed method generates well-positioned centerlines, exhibiting lower number of missing branches and is more robust in the presence of  minor imperfections of the object segmentation mask. 
Compared to a state-of-the-art traditional minimal path approach, our method improves patient-level success rate of centerline extraction from 54.3\% to 88.8\% according to independent human expert review.

\end{abstract}

\section{Introduction}

Centerline, or skeleton, provides a concise representation of the object topology. An ideal centerline extraction algorithm generates centerline points close enough to ``centers'' of the object cross-sectionally, captures all ``branches'', and has no false positive spurious branches. 
\blfootnote{* Zhihui Guo and Junjie Bai -- equal contribution}

Many semi-automated and automated approaches exist for centerline extraction. Morphological thinning and erosion based methods are popular in road centerline extraction \cite{jin2016robust}. However, centerlines extracted by these methods often come along with spurious branches and usually need ad-hoc post pruning. 

In contrast, minimal path based centerline extraction methods guarantee a more structured centerline output by requiring explicitly specified endpoints. A \emph{centerline distance map} (also called cost image) is generated from an object segmentation mask by methods such as Euclidean distance transform, which assigns smaller values to voxels closer to centerline and larger values to voxels farther away. A minimal path between the endpoints in the centerline distance map thus corresponds to the object centerline. 

%Besides, the performance of these methods also depend upon the underlying cost metrics. A common choice of cost metrics is centerline distance map, which reflects how close each voxel is to the desired centerline. The voxels closer to centerline location are assigned a lower value in the distance map, while the other voxels are given a higher value. Thus, a path that minimizes cost along the route on the distance map between paired end points is found. 

Minimal path algorithms are widely used in blood vessel centerline extraction \cite{metz2009coronary,jin2016robust,gulsun2016coronary,mirikharaji2017globally,zheng2013robust}. In \cite{metz2009coronary}, Metz \textit{et al.} adopted vesselness and region statistics as cost metrics and manually specified each branch endpoint. G{\"u}ls{\"u}n \textit{et al.} \cite{gulsun2016coronary} used human selected features to compute flow field as cost image. Mirikharaji \textit{et al.} \cite{mirikharaji2017globally} integrated a predefined tree topology and tubularity scores to get minimal paths. Zheng \textit{et al.} \cite{zheng2013robust} used a machine learning based vesselness algorithm to generate the cost image. 
These methods require either human crafted features/priors or manual specification of branch endpoints. 
%These methods require human intervention to either define features and priors or detect branch end points.

Convolutional neural networks (CNN) have been prevalent in medical image analysis recently and achieved great success.
% In a CNN model, features are extracted by convolutional operations and passed through multi-layer architecture during forward propagation. Weights of the model are updated by gradient-based optimization during backward propagation. 
There are three advantages of CNN-based methods over traditional methods. First, multi-layer CNNs have enough capacity to learn complex functions that cannot be described by simple models. Second, CNNs do not require humans to select features and support end-to-end training. Third, a single CNN model has the ability to handle multiple tasks. 
%It is promising to use a fully convolutional network (FCN) to compute centerline distance maps and branch end points simultaneously. 
%To the best of our knowledge, this is the first time that a single network predicts both distance maps and branch end points in minimal path algorithms for tree-like objects.

Coronary computed tomography angiography (CCTA) is a noninvasive technique widely used in clinical practice for coronary artery disease detection. Given a coronary artery segmentation mask, extracting its centerline is a prerequisite step for automatic stenosis grading, calcium evaluation, plaque evaluation and visualization~\cite{xiong2017comprehensive,kiricsli2013standardized}.
Extracting coronary artery centerline from a segmentation mask faces multiple notable challenges. First, multiple branches, usually more than a dozen, with large intra-subject and inter-subject variations of length, thickness, and shape are presented, forming a complex tree structure. Detecting all branches without false positive is quite challenging.
%besides the main branch, multiple side branches are also presented, which makes accurate detection of all branch ending points quite difficult. 
Second, branch diameter changes significantly from proximal to distal portion of coronary artery. The proximal end can be several times wider than the distal end. Third, tortuous course of vessel branches hinders the performance of a minimal path based algorithm, by which straight paths are inherently preferred. Fourth, imperfections of segmentation masks such as brief touching of two nearby branches could lead to incorrect bifurcation in centerline.

To address these challenges, we propose a two-head multi-task FCN which simultaneously generates a locally normalized distance map and a list of branch endpoints (Fig.~\ref{fig:workflow}). One head of the multi-task FCN outputs a normalized centerline distance map that is scale-invariant and robust to image segmentation imperfections. Log-transform and attention mechanism are also incorporated to increase model sensitivity. The other head of the FCN automatically detects the sparsely distributed endpoints of the object skeleton with high accuracy. The resulting distance map and endpoint list are fed into a minimal path extractor which gives the final centerline extraction results. 

%Coronary computed tomography angiography (CTA) is a noninvasive technique that has been widely used in clinical practice for coronary artery disease (CAD) detection \cite{norgaard2014diagnostic}. However, for tasks like blood flow simulation, automated calcium evaluation and stenosis grading, centerline extraction serves as the first step\cite{xiong2017comprehensive,kiricsli2013standardized}. Centerlines are also regarded as a straightforward but powerful visualization tool to interpret vessel branch growth and topology. 
\section{Method}

%%%%%%%%%%%%%%%%%%%%%%%%%%%%%%%%%%% Figure %%%%%%%%%%%%%%%%%%%%%%%%%%%%%%%%%%%%%
\begin{figure}[t]
\centering
\includegraphics[width=0.9\textwidth]{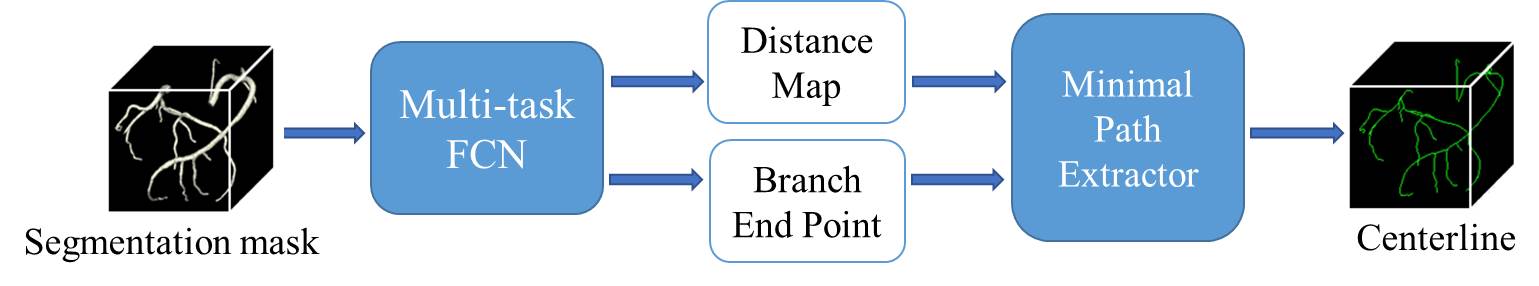}
\caption{Schematic workflow of DeepCenterline} \label{workflow}
\label{fig:workflow}
\end{figure}
%%%%%%%%%%%%%%%%%%%%%%%%%%%%%%%%%%%%%%%%%%%%%%%%%%%%%%%%%%%%%%%%%%%%%%%%%%%
The proposed method consists of two main steps (Fig.~\ref{fig:workflow}): a multi-task FCN computing a locally normalized centerline distance map and a list of endpoints simultaneously, and a minimal cost path extractor taking the output of the FCN to generate a set of paths as centerline. 
%Modules used in the multi-task FCN and the way of training on the two tasks of centerline distance map computation and ending points detection are discussed in section \ref{subsec:fcn}. The minimal cost path extractor that generates a well-structured centerline of coronary artery is displayed in section \ref{subsec:path}.

%We propose a novel framework for coronary centerline extraction (Fig. 1). From coronary CTA scans,  coronary arteries including side branches are first segmented by a variant of UNet \cite{ronneberger2015u} to obtain binary coronary arterial masks. The binary masks are further refined by humans to fix unexpected breaks of vessels. The FCN model then takes the refined coronary masks as input and is trained by two label maps for centerline distance metric and branch end point respectively. A minimal path extractor is applied utilizing predictions of distance map and end points to get centerlines for entire branches. Note that this work only focuses on the process from coronary mask to extract coronary centerlines.
\subsection{Multi-task FCN architecture}
\label{subsec:fcn}
%%%%%%%%%%%%%%%%%%%%%%%%%%%%%%%%%%% Figure %%%%%%%%%%%%%%%%%%%%%%%%%%%%%%%%%%%%%
\begin{figure}[t]
\centering
\includegraphics[width=\textwidth]{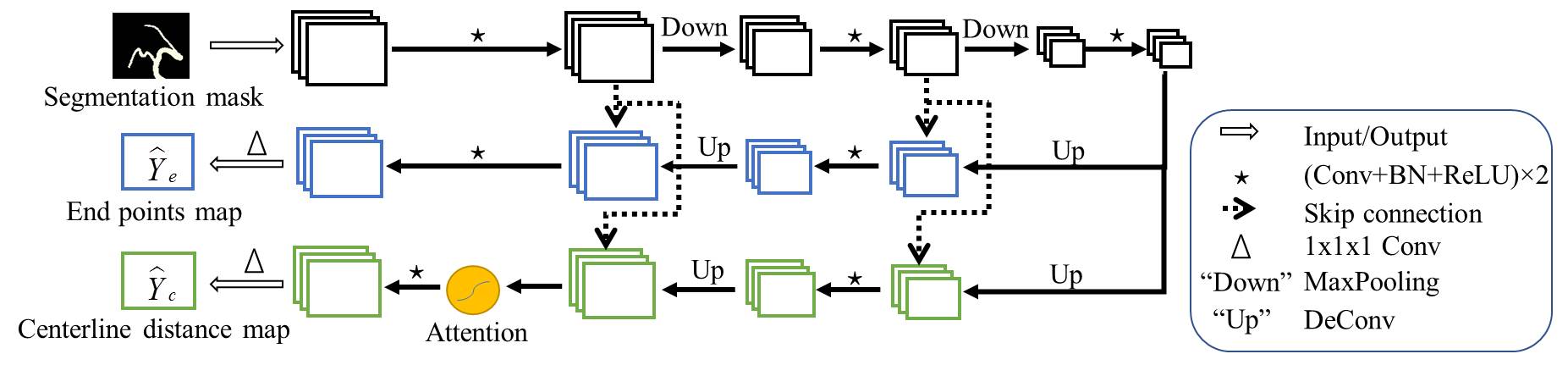}
\caption{The proposed multi-task FCN architecture. The input is 3D segmentation mask volume. 
The two tasks, centerline distance map and endpoint confidence map computation, share the same encoder path and have separate decoder paths. 
Skip-connections are added among features of same scale to facilitate good use of information. An attention module is added for the centerline distance map task to further boost accuracy. 
%The input is 3D segmentation mask volume. Deep features are extracted by sequential convolutions, batch normalization, ReLU activation function and max pooling. We employ upsampling to perform end-to-end training. Multi-scale features are incorporated by skip connections. Predictions for centerline distance map and branch end points are outputted from the model.
} 
\label{fig:fcn}
\end{figure}
%%%%%%%%%%%%%%%%%%%%%%%%%%%%%%%%%%%%%%%%%%%%%%%%%%%%%%%%%%%%%%%%%%%%%%%%%%%
The multi-task FCN accomplishes two tasks: computing a normalized centerline distance map and detecting branch endpoints. As shown in Fig.~\ref{fig:fcn}, the two tasks share the same encoder layers consisting of convolution (Conv), batch normalization (BN), ReLU activation function and max pooling (shown as `Down' in Fig.~\ref{fig:fcn}) operations. The two tasks then have different decoder layers tailored for each task, consisting of conv, BN, ReLU, and upsampling operations. Skip connections are applied at the same scale between encoder and decoder layers to make effective use of both high-level and low-level features similar to~\cite{ronneberger2015u}. 
Suppose a volumetric segmentation mask is $I^{X\times Y \times Z}$, and $\mathcal{I} = \{(x,y,z)| x\in \{1,\ldots,X\}, y \in \{1,\ldots,Y\}, z\in \{1,\ldots,Z\}\}$ denotes the set of all voxel locations in the image.

%Our network is consisted of encoder, decoder and attention modules, as shown in Fig.\ref{fig:fcn}. An encoder module extracts deep features of the input by a set of convolution (Conv), batch normalization (BN), activation and max pooling (maxPool) operations. An decoder module then takes deep features as input and follows a set of Conv, BN, activation and upsampling operations to complete the basic requirement of end-to-end training. In Fig. 2 (a), both tasks share the same encoder to reduce number of parameters. Skip connections are applied to concatenate features with the same scale in encoder and decoder to use multi-scale features as in \cite{ronneberger2015u}.

%Label maps representing centerline distance map and branch end points are described below.
\subsubsection{Centerline distance map}
\label{dist}
A centerline distance map is defined as an image whose voxel intensity shows how close each voxel is to the nearest centerline point.
Due to the large variations in branch radius (coronary artery proximal radius can be five times bigger than the distal radius), 
a straightforward Euclidean distance transform computation generates centerline distance map with largely variable range of values at different sections of the branch. To obtain a centerline consistently well-positioned in the ``center'' from beginning to end requires tricky balancing of cost image contrast between thick and thin sections. 
% * <zhihui-guo@uiowa.edu> 2018-12-06T22:10:42.036Z:
%
% ^.
%The straightforward way to compute a centerline distance map given a segmentation mask is to compute the Euclidean distance transform. However, the distance map obtained in this way has different value ranges for thick vessels and thin vessels. For example, the proximal coronary artery may have diameter that is \~5 times wider than the distal side. Thus, the local contrast of the distance map is poor at some portion of the vessel. Applying minimal cost path algorithm on distance map with such a large range/contrast variation makes the resulting centerline to take straight off-center ``shortcut'' at tortuous vessel part due to the poor distance map contrast at these regions. 

%The proposed FCN generates a locally normalized centerline distance map to achieve the desired scale-invariance. 
To achieve the desired scale-invariance property, we propose to use FCN to generate a \emph{locally normalized} centerline distance map.
More specifically, during training, a local cross-sectional view of the segmentation mask perpendicular to the centerline tangent direction at each centerline point is obtained. Suppose the set of all foreground voxels in the cross-sectional view is $\mathcal{S}$. Then the locally normalized distance map value for voxel index $i=(x,y,z) \in \mathcal{S}$ is computed as 
\begin{align}
d_i = \frac{d_i^{Euc}}{\max_{i \in \mathcal{S}}{d_i^{Euc}}}
\label{eq:normalizedDist}
\end{align}
where $d_i^{Euc}$ is the Euclidean distance of voxel $i$ to the centerline point on the view. 
%Note this distance computation is only required for generating training references and is not needed during testing stage. 
To further highlight contrast at portions closer to centerline, log-transform is applied to generate the reference centerline distance map
\begin{align}
\mathbf{Y_c} = \log(\mathbf{D_c} + \delta)
\label{eq:Y_c}
\end{align}
where $\mathbf{D_c} = \{d_i| i \in \mathcal{I}\}$ is the locally normalized centerline distance map throughout the whole segmentation mask image $I$ and $\delta$ is a small positive constant to avoid numerical issues. Note that the distance computation in Eq.\eqref{eq:normalizedDist} and Eq.\eqref{eq:Y_c} is only carried out when generating training reference standards. During testing phase, the FCN will directly predict centerline distance map $\mathbf{\widehat{Y}_c}$. 
%To avoid shortcuts and false bifurcations that traditional methods are apt to happen, we aim to have a distance map that has two properties. Firstly, the distance map should be scale-invariant. Coronary arterial radius changes significantly from proximal to distal, while balancing the uneven radius is nontrivial. Shortcuts in minimal path is majorly caused by the underlying cost metrics being dominated by either side of the radius. To achieve scale-invariant, we normalize euclidean distances of vessel voxels to centerline points on the same cross-sectional plane to [0, 1]. Secondly, the distance map should have strong and intense values at locations close to centerline for the neural network to notice. Log-transform is applied to the normalized distance map to make the center area highly distinguishable to the rest. Besides, attention mechanism as described above is also enforced to focus on center area. 
%\begin{equation}
%\label{eq:cl}
%Y_{c}=log(D_{c}+\delta )
%\end{equation}
%$D_{c}$ is the normalized Euclidean distance matrix and $\delta $ is set as $e^{-10}$.
\vspace{-10pt}
\subsubsection{Spatial-wise and channel-wise attention for centerline distance map}
%%%%%%%%%%%%%%%%%%%%%%%%%%%%%%%%%%% Figure %%%%%%%%%%%%%%%%%%%%%%%%%%%%%%%%%%%%%
\begin{figure}[t]
\centering
\includegraphics[width=0.8\textwidth]{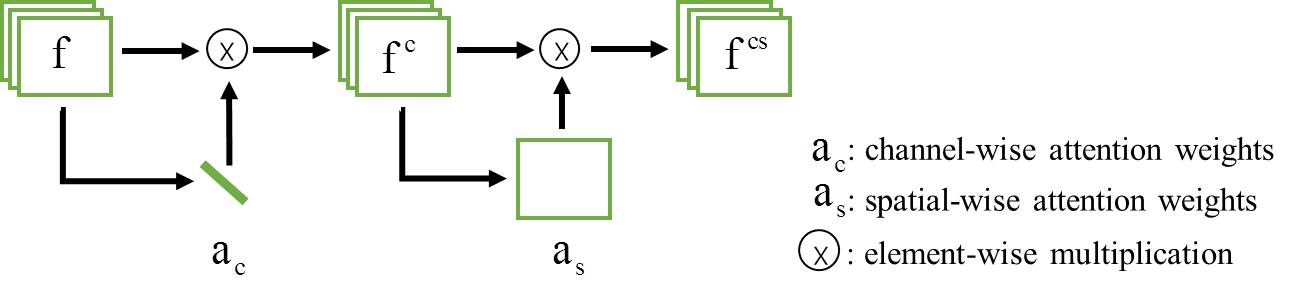}
\caption{Spatial-wise and channel-wise attention} 
\label{fig:fcn-1}
\end{figure}
%%%%%%%%%%%%%%%%%%%%%%%%%%%%%%%%%%%%%%%%%%%%%%%%%%%%%%%%%%%%%%%%%%%%%%%%%%%
Traditionally the convolutional features at different spatial locations and channels are treated equally by the following layers. 
%However, centerline distance map needs to have sufficient distinguishing power at a very narrow region around centerline point. 
However, centerline extraction is inherently a localized task. Specifically, the narrow region surrounding the underlying centerline requires  most attention for best discriminating contrast. 
Thus, a spatial-wise attention module is proposed to weight feature maps at different spatial locations. Similarly, different channels of the feature maps can highlight different regions (some channels may focus more around centerline, while other channels may focus more on object mask boundary, etc). It also makes sense to add a channel-wise attention to weight different channels accordingly. Fig. \ref{fig:fcn-1} shows the proposed spatial-wise and channel-wise attention module, inspired by~\cite{zhang2018progressive}. The feature map $\mathbf{f}$ is first weighted by channel-wise attention, generating $\mathbf{f^c}$, and then weighted by the spatial attention, generating $\mathbf{f^{cs}}$. 

%Suppose the feature map before the attention module is $\mathbf{f}\in \mathbb{R}^{I\times C}$, where image size $I$ is $X\times Y \times Z$ along three axis respectively and number of channel is $C$. The spatial and channel-wise attention modules are added similar to ~\cite{zhang2018progressive}. As shown in Fig.\ref{fig:fcn}, the feature map $\mathbf{f}$ is weighted by channel-wise attention first, generating $\mathbf{f^c}$, and then weighted by the spatial attention, generating the final feature maps $\mathbf{f^{cs}}$.  

Channel-wise attention weights different channels by vector $\mathbf{a_c} \in \mathbb{R}^C$, where $\sum_{i=1}^C \mathbf{a_c}(i) = 1$. To obtain this weighting vector, average pooling is first applied on each channel of the feature map to obtain a summarized channel feature vector $\mathbf{v} \in \mathbb{R}^C$. Then a convolutional layer and a ReLU nonlinearity $\sigma(\cdot)$ are added to obtain the raw attention weights $\mathbf{u} \in \mathbb{R}^C$. In Eq.\eqref{eq:channelRaw}, $*$ is the convolution operator. $\mathbf{W_c}$ is the convolutional kernel and $\mathbf{b_c}$ is the bias vector. A softmax function is applied on the raw attention weights $\mathbf{u}$ to obtain the final channel-wise attention vector $\mathbf{a_c}$, as shown in Eq.\eqref{eq:channelSoftmax}. 
\begin{align}
\mathbf{u} &= \sigma(\mathbf{W_c}*\mathbf{v} + \mathbf{b_c}) \label{eq:channelRaw}\\
\mathbf{a_c}(i) &= \frac{e^{\mathbf{u}(i)}}{\sum_{c=1}^{C}e^{\mathbf{u}(c)}}, \quad i \in \{1,\cdots,C\} \label{eq:channelSoftmax}
\end{align}
To apply the channel-wise attention weights, the input feature map at channel $i$ is multiplied by attention weight $\mathbf{a_c}(i)$ to obtain channel-weighted feature map
% suppose the input feature map at channel $i$ is $\mathbf{f}_i$, then we obtain $\mathbf{f}_i^{\mathbf{c}}$ by $\mathbf{f}_i$ multiplying the corresponding weight $\mathbf{a_c}(i)$ (Eq.\eqref{eq:channelFeature}). 
%
\begin{align}
\mathbf{f}_i^{\mathbf{c}} &=  \mathbf{a_c}(i) \cdot \mathbf{f}_i, \quad i \in \{1,\cdots,C\} \label{eq:channelFeature} 
\end{align}

Spatial-wise attention weight matrix $\mathbf{a}_s \in \mathbb{R}^I$ is obtained similar to the channel-wise attention. The raw spatial attention map $\mathbf{q}$ is computed by applying a $1\times 1 \times 1$ convolutional layer $\mathbf{W_s}$ with bias $\mathbf{b_s}$ and a ReLU nonlinearity $\sigma(\cdot)$ to $\mathbf{f^c}$ (Eq.\eqref{eq:spatialRaw}). Then a softmax function is used to obtain the final spatial attention weights $\mathbf{a_s}$, where $\sum_{i\in \mathcal{I}}\mathbf{a_s}(i) = 1$. The spatial attention weight is applied by multiplying $\mathbf{a_s}(i)$ with features $\mathbf{f}_i^{\mathbf{c}}$ at location $i$ (Eq.\eqref{eq:spatialFeature}). 
%Suppose features at location $i\in \mathcal{I}$ is $\mathbf{f^c}_i$, then the spatial attention weight $\mathbf{a_s}(i)$ is multiplied with it to obtain the weighted output feature value (Eq.\eqref{eq:spatialFeature}). 
%The spatial attention weight is applied to the feature map after channel-wise attention by a Hadamard matrix product operation $\odot$. This product operation multiplies feature map values with location-specific weights.
%
\begin{align}
\mathbf{q} & = \sigma(\mathbf{W_s} * \mathbf{f_c} + \mathbf{b_s}) \label{eq:spatialRaw} \\
\mathbf{a_s}(i) &= \frac{e^{\mathbf{q}(i)}}{\sum_{j\in \mathcal{I}} e^{\mathbf{q}(j)} } , \quad i \in \mathcal{I} \label{eq:spatialSoftmax} \\
%\mathbf{f^{cs}} &= \mathbf{a_s} \odot \mathbf{f^c} \label{eq:spatialFeature}
\mathbf{f}_i^{\mathbf{cs}} &= \mathbf{a_s}(i) \cdot \mathbf{f}_i^{\mathbf{c}}, \quad i \in \mathcal{I} \label{eq:spatialFeature}
\end{align}

\subsubsection{Branch endpoint detection}
Different from centerline distance map which consists of continuous values inside the whole segmentation mask, branch endpoints are just a few isolated points. Directly predicting these points using a voxel-wise classification or segmentation framework is not feasible due to the extreme class imbalance. To tackle the class imbalance problem, a voxel-wise endpoint confidence map is generated by constructing a Gaussian distribution around each endpoint to occupy a certain area spatially. 
The FCN is then trained to predict the endpoint confidence map, which has a more balanced ratio between nonzero and zero voxels. 
%In this way, we relief the class imbalance problem by increasing the number of nonzero-valued voxels in the confidence map. The FCN is then trained to predict the end point confidence map. 

Specifically, a spatial Gaussian field is generated around each endpoint
\begin{align}
\mathbf{{Y}_e}(i) = \frac{1}{\sqrt{2\pi \Delta}} e^{-\frac{\mathbf{D_e}(i)^2}{\Delta^2}}, \quad i \in \mathcal{I}
\label{eq:Y_e}
\end{align}
where $\mathbf{D_e}(i)$ is the geodesic distance from voxel $i$ to the nearest branch endpoint inside the segmentation mask. $\Delta$ controls the scale of the Gaussian field. 

In testing phase, the predicted endpoint confidence map is thresholded by half of the maxium possible value, i.e., $0.5/\sqrt{2\pi \Delta}$. The centroid of each connected component is then returned as branch endpoints. 

%During the testing, after the prediction of the end point confidence map is generated by the FCN, a simple connected component analysis is conducted after thresholding the confidence map by a threshold of $0.5/\sqrt{2\pi \Delta}$. The centroid of each connected component is then assigned as the detected branch end points. 

% Different from fully connected regions or vessel centerlines which are composed of continuous points, branch end points are extremely sparse. Direct prediction of branch end points is impossible. This is similar to the situation in \cite{lalonde2018clusternet}, where the task was to detect discrete tiny objects from a large background. To make them recognizable by FCN, enlarging the impact region for each end point is necessary. Here we created a Gaussian field for each end point. 
% \begin{equation}
% \label{eq:end}
% Y_{e}=\frac{1}{\sqrt{2\pi \sigma }}e^{-\frac{D_{e}^{2}}{\sigma ^{2}}}
% \end{equation}
% In Eq. \ref{eq:end}, $D_{e}$ is the distance matrix representing to the nearest end point inside the vessel mask. $\sigma $ controls the scale of the effected region, which is set as 3mm in this work.

\subsubsection{Loss function}
The loss function shown in Eq. \eqref{eq:loss} consists of two terms, one for the centerline distance map prediction and the other for the branch endpoint detection. We enforce the loss function to only account for regions inside the segmentation mask. Suppose the segmentation mask is $\mathbf{\Lambda} \in \{0,1\}^{X\times Y \times Z}$. $\mathbf{\Lambda}(x,y,z)$ is 1 for every voxel $(x,y,z)$ inside the segmentation mask, and 0 otherwise.
%, $\mathbf{\Lambda}(x,y,z)$ is 1. Otherwise, $\mathbf{\Lambda}(x,y,z)$ is 0. 
%Let $\odot$ denote the Hadamard matrix product operation, which is element-wise multiplication of two matrices. Then the loss function is listed as below: 
%
\begin{align}
L = \gamma  \| \Lambda \odot (\mathbf{Y_c} - \widehat{\mathbf{Y}}_\mathbf{c}) \|^2 + (1-\gamma)  \| \Lambda \odot (\mathbf{Y_e} - \widehat{\mathbf{Y}}_\mathbf{e}) \|^2
\label{eq:loss}
\end{align}
In Eq. \eqref{eq:loss}, $\mathbf{Y_c}$ and $\widehat{\mathbf{Y}}_\mathbf{c}$ are the reference standard and the predicted centerline distance map. $\mathbf{Y_e}$ and $\widehat{\mathbf{Y}}_\mathbf{e}$ are the reference standard and the predicted endpoint confidence map. $\odot$ denotes the Hadamard matrix product operation, which is element-wise multiplication of two matrices. $\gamma $ is a weighting factor that balances losses of centerline distance map and branch endpoint confidence map.

\subsection{Minimal path extraction}
\label{subsec:path}
Given a root point, a list of branch endpoints, and the underlying centerline distance map, a minimal path algorithm is used to extract the centerline of a tree-structured object. 

We construct an undirected graph $G=(\mathcal{V},\mathcal{E})$, where set $\mathcal{V}$ contains all vertices corresponding to voxels in the segmentation mask and set $\mathcal{E}$ includes all edges connecting two neighboring vertices in set $\mathcal{V}$ under a 26-neighborhood setting. On each vertex $v_i$, weight $w_{v_i}=\exp(\widehat{\mathbf{Y}}_\mathbf{c}(i))$ is set according to the centerline distance map.
%the centerline distance map value at this voxel $\exp(\mathbf{Y_d}(i))$ is assigned by the vertex weight $w_{v_i}$. 
Note that each vertex carries a weight that is smaller when the corresponding voxel is closer to the centerline location, and larger when it is farther away from centerline. Given a starting vertex $s\in \mathcal{V}$ and an ending vertex $t\in \mathcal{V}$, a minimal path between the two is defined as $\mathbf{p} = (p_1, p_2, \ldots, p_K), p_k \in \mathcal{V}, k=\{1,2,\ldots,K\}$ such that (1). $p_1=s$, $p_K = t$; (2). every two neighboring vertices in the path is connected by an edge; (3). the sum of vertex weights along this path is minimized (Eq. \eqref{eq:minPath}). 
\begin{align}
\mathbf{p} = \argmin_{(p_{k}, p_{k+1}) \in \mathcal{E}}\sum_{k=1}^{K} w_{p_k}
\label{eq:minPath}
\end{align}
Such a minimal path from $s$ to $t$ corresponds to the desired centerline between the two points. To extract the centerline of a tree-structured object such as a coronary artery tree, one root point usually correspond to multiple branch endpoints $t_1, t_2, \ldots, t_T$. In this case, the minimal path between root point $s$ and each endpoint $t_i, i\in \{1,2,\ldots,T\}$ is computed respectively. Then we trace each path from the end to the start point sequentially. Once the current path intersects with some previously traced paths, it is merged into the previously traced paths.
%stop the tracing of the current path, which effectively merges it with the previously traced path until the root point. 
%This merging does not compromise the minimal path property of current path since a minimal path segment between two points for one path is also a minimal path for all paths passing the two points. 
The centerline points are finally smoothed by an iterative mean filtering in a small window for smoother appearance.

\section{Experiments and Results}
\subsection{Experimental design}
To evaluate the proposed method, 620 volumetric coronary CTA scans of 620 patients are used. The image spacing is first normalized to $0.4\times 0.4\times 0.4mm^{3}$. Coronary arteries and ascending aorta are segmented by a semi-automatic software with manual review and editing. The segmentation masks of coronary arteries and ascending aorta serve as input to the experiment. Since coronary artery originates from ascending aorta, the root points of each coronary vessel tree are readily available as the artery voxels connected to aorta. To simplify notation, we use \emph{CL} as a shorthand for `centerline'. 

Manual annotations of centerline are hard to obtain due to the complex 3D structure of vessels and the single-pixel-wide requirement. Thus, during training, centerlines extracted by a state-of-the-art traditional method (called \emph{baseline}) serve as the training reference truth for DeepCL. During testing, the degree of matching between DeepCL and baseline is first studied as a sanity check. Then various metrics requiring no ``truth'' centerline such as centerline to segmentation mask Hausdorff distance, and independent human expert review, are utilized to evaluate DeepCL and baseline method. 
\vspace{-10pt}
\subsubsection{Baseline}
% \xxx{Should we include more descriptions about the baseline methods,like
% \begin{equation}
% \mathbf{D^{sig} = \sum_{i=1}^3 1 / (1+ \exp(-(\mathbf{D^{Euc}} - \beta_i)/ \alpha_i))}
% %D_{s }=Sigmoid (D_{e}, \Theta )
% \end{equation}
% where $D_{e}$ is the Euclidean distance matrix and $D_{s }=\{d_{i,j,k}|d_{i,j,k}\in [0,1]\}$ is the transformed distance map. Parameters $\Theta $ adjust $D_{e}$ by scale and shift operations, the values of which were optimized on the whole dataset. 
% }

The baseline method is also a minimal path approach. However, both branch endpoints and centerline distance map are computed by traditional methods. The centerline distance map $\mathbf{D^{Sig}}$ is derived from the Euclidean distance map $\mathbf{D^{Euc}}$ of the segmentation mask by summing three sigmoid functions to highlight the contrast in regions close to the centerline area. 
\begin{equation}
\mathbf{D^{Sig}} = \sum_{i=1}^3 \frac{1}{1+ \exp(-\frac{\mathbf{D^{Euc}} - \beta_i}{ \alpha_i})}
\label{eq:baseline}
%D_{s }=Sigmoid (D_{e}, \Theta )
\end{equation}
Three pairs of parameters ($\alpha_i, \beta_i$), each controlling the width and level of a contrast window, are tuned to enhance central contrast for vessel segments with large, medium, and small diameters respectively. Summing of these three sigmoid functions results in a relatively good contrast around centerline area throughout the whole vessel tree. 
%$\mathbf{D^{sig} = \sum_{i=1}^3 1 / (1+ \exp(-(\mathbf{D^{Euc}} - \beta_i)/ \alpha_i))}$ 
The branch endpoints are detected as local maxima of the arrival time by a breadth first search, starting from the root point at the junction of aorta and artery to each voxel throughout the artery segmentation mask. All related parameters are tuned on a different dataset. 
%As we will see from the experiment result, this baseline is a strong baseline achieving good branch-wise accuracy. 
\vspace{-10pt}
\subsubsection{DeepCenterline}
We randomly divided 620 scans into three dataset: 200 scans for training, 20 scans for validation and 400 scans for testing. On the training set, centerlines extracted by the baseline method are used as the reference truth. Although generated by a strong baseline method, the reference truth still contains errors such as missing branches, wrong bifurcations in case of imperfect segmentation mask, etc. The locally normalized centerline distance map and the branch endpoint confidence map are generated based on the reference truth. The parameters of the proposed method are tuned based on the validation set. The tuned model is applied to the 400 testing scans to evaluate the performance. 

The input patch size is $64\times 64\times 64$ voxels. The standard deviation of Gaussian field $\Delta$ in Eq. \eqref{eq:Y_e} is set to 3 mm. The loss weighting coefficient $\gamma$ in Eq. \eqref{eq:loss} is 0.5. 
%We avoid to train patches containing no vessel voxels. 
Our multi-task FCN network is optimized by stochastic gradient descent, with batch size of 3. Total number of epochs is 20. The initial learning rate is $10^{-2}$, which is divided by a factor of 2 every 5 epochs.

%Note that errors generated from baseline method are also included in training set. During training phase, labels for centerline distance map and branch end points were generated taking baseline results as ground truth. During testing phase, distance map and branch end points were predicted simultaneously on test set with trained FCN model. 

\subsubsection{Evaluation metrics}
Several evaluation metrics based on either objective metrics or independent human expert review are used for a thorough comparison of the performance of baseline method and DeepCL on the test set.
\begin{enumerate}[i]
  \item Mean centerline to centerline distance. The mean absolute distance from centerline A to centerline B is defined as the mean of the absolute distance to the nearest point on B for every point on A. 
  \item Coverage percentage. A point on centerline A is covered by centerline B if the closest point on B is within half a voxel (0.2 mm). 
  \item Number of missing endpoints.  The number of endpoints not found by automated algorithm is counted manually in patient-level as well as branch-level. 
  \item Number of scans with wrong bifurcations. This usually happens when two branches are spatially close or even briefly joined at a certain section. The centerline could wrongly consider this brief joining as a bifurcation. 
%We collected scans that had significant false positive bifurcations. This often happens when two branches are partially close and the centerline for one branch jumps to the other branch.
  \item Average patient-level centerline length. The patient-level centerline length is computed as the sum of lengths of each centerline segment. 
  %The same segment shared by multiple branches is only counted once. 
  In general, the less straight ``shortcut'' centerline takes, and/or the more branch endpoints are detected, the longer centerline will be. 
  %We also considered the average length of centerlines. Principally, the less shortcuts there exist, and/or the more branch end points are detected, the longer centerlines will be. 
  \item Hausdorff distance. Hausdorff distance is defined as the maximum of distances from every artery segmentation mask voxel to the closest centerline point. Hausdorff distance shows how close each segmentation voxel is being covered by the extracted centerline. 
  %The farthest distance from coronary mask voxels to the corresponding centerline was explored for each scan. We studied the distribution of Hausdorff distance distribution for 400 scans, and observed which method was further deviated from center area.
  \item Overall success rate. A centerline extraction is called fully successful when an expert reviews the centerline and determines that the centerline covers all branches sufficiently, has no spurious false positive branch, no wrong bifurcation, and no obvious deviation from the center throughout all sections. 
  %As a straightforward indicator, the success rate tells the proportion of scans for which their centerlines are fully accepted by experts.
\end{enumerate}

\subsection{Results}
\begin{figure}[t]
\includegraphics[width=\textwidth]{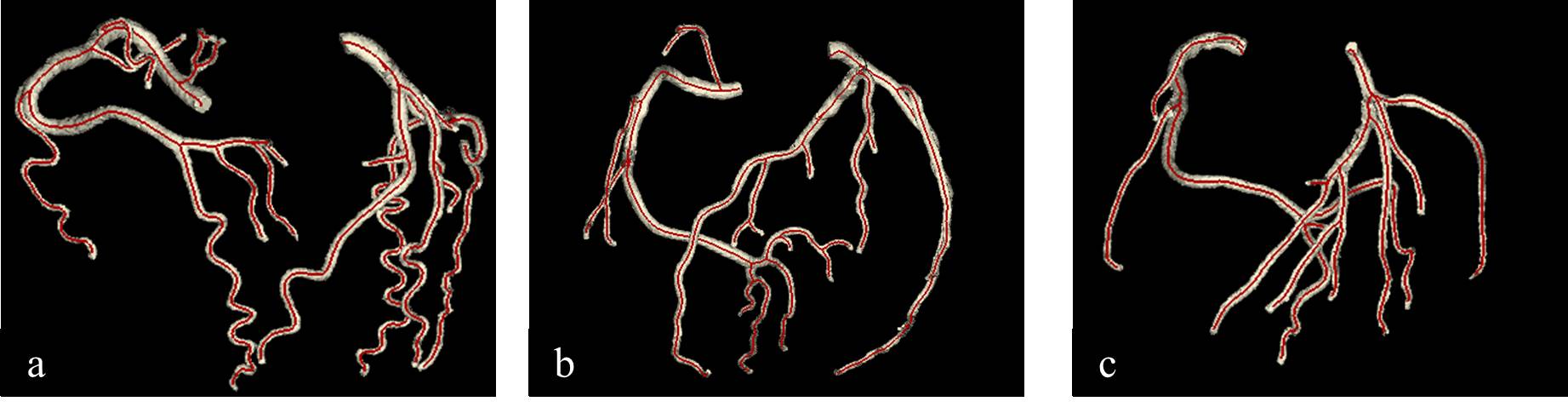}
\caption{Examle coronary artery with: (a) very tortuous course; (b) small branches; (c) two spatially close branches. Red lines are centerlines generated by DeepCL.} 
\label{fig:whole}
\end{figure}
Fig. \ref{fig:whole} displays three examples of coronary artery segmentation masks overlaid with centerlines extracted by DeepCL. For each coronary artery, radius changes substantially from the proximal to the distal side. Different coronary arteries have large variations of vessel curvature, shape and branch topology. Despite all these difficulties, our method is able to extract well-positioned centerline for all branches without false positive branches. 

%Due to the lack of manually annotated reference, we first verify the degree of matching between DeepCenterline and the baseline method as a sanity check. 
%Objective measurements requiring no reference standard and subjective evaluation from independent human expert are then used to examine the differences between DeepCenterline and the baseline in detail. 
%Lastly, the effectiveness of proposed attention module is examined. To simplify notation, we use \emph{CL} as a shorthand for `centerline'. 

\subsubsection{Matching to baseline}
\begin{table}
\renewcommand{\arraystretch}{1.2} % increase row height in table, default is 1
\setlength{\tabcolsep}{10pt} % increase column width in table, default is 6pt
\caption{Degree of matching between DeepCL and baseline method. A$\rightarrow$B measures the distance from one point on centerline A to nearest point on centerline B.}\label{tab:match}
\centering
\begin{tabular} {c|c|c}
\hline 
& baseline $\rightarrow$  DeepCL & DeepCL $\rightarrow$  baseline \\ \hhline{=|=|=}
mean distance (mm) & 0.066$\pm$0.053  & 0.068$\pm$0.014 \\ \hline 
being covered by (\%) & 99.7$\pm$0.7 & 99.5$\pm$0.6 \\ \hline 
% \begin{tabular}{|c|c|c|}
% \hline
%                                                         & \begin{tabular}[c]{@{}c@{}}Baseline $\rightarrow$  DeepCenterline\end{tabular} & \begin{tabular}[c]{@{}c@{}}DeepCenterline $\rightarrow$  Baseline\end{tabular} \\ \hline
% \begin{tabular}[c]{@{}c@{}}Mean Distance (mm)\end{tabular} & 0.066$\pm$0.053                                                                  & 0.068$\pm$0.014                                                                  \\ \hline
% \begin{tabular}[c]{@{}c@{}}Coverage (\%)\end{tabular} & 99.7$\pm$0.7                                                                     & 99.5$\pm$0.6                                                                     \\ \hline
\end{tabular}
\end{table}

Table \ref{tab:match} shows the degree of matching between DeepCL and baseline method. The mean centerline distance and ``being covered by'' percentage both shows how close/well one centerline is being covered by the other centerline. The low mean distance value and the very high coverage percentage on both direction (baseline to DeepCL and DeepCL to baseline) show that the two methods are in good alignment in general. However, a larger portion of baseline centerline points are being covered by DeepCL centerline (99.7\%) by a smaller distance (0.066 mm) than the other way around (99.5\%, 0.068mm). 
%Moreover, baseline centerline is also covered more closely covered by DeepCL centerline (0.066 mm) than the DeepCL being covered by baseline (0.068 mm). 
This implies DeepCL provides slightly better coverage than baseline, which will be assessed in detail in the following analysis. 

Notably, both DeepCL and baseline generate no spurious false positive branches in the extracted centerline according to visual inspection. 
%In fact, a more careful analysis in the next section reveals that DeepCL does provide a better coverage due to more branches, more accurate bifurcation, and longer centerlines being found, etc.
%is measured from centerline A $\rightarrow$ B by measuring the distance to nearest B point from every point on A. This demonstrates how well centerline B ``covers'' centerline A. The ``being covered by'' percentage shows the portion of points on centerline $A$ for which a poing on centerline $B$ lies within the distance threshold of 0.2mm (half a voxel). 

\subsubsection{Performance difference}
\begin{table}[t]
\renewcommand{\arraystretch}{1.2} % increase row height in table, default is 1
\setlength{\tabcolsep}{8pt} % increase column width in table, default is 6pt
\caption{Difference in performance of DeepCL and baseline.}\label{tab:comp}
\centering
\begin{tabular}{c|l||cc|cc}
\hline
\multicolumn{2}{c||}{\multirow{2}{*}{metrics}}                                                       & \multicolumn{2}{c|}{raw number \#/\#} & \multicolumn{2}{c}{ratio \%} \\ \cline{3-6} 
\multicolumn{2}{c||}{}                                                                        & baseline     & DeepCL      & baseline    & DeepCL    \\ \hhline{=|=#==|==}
\multirow{2}{*}{\begin{tabular}[c]{@{}c@{}}missing \\ endpoint\end{tabular}} & patient-level & 170/400      & \textbf{34/400}      & 42.5\%      & \textbf{8.5\% }    \\ \cline{2-6} 
                                                                              & branch-level  & 233/6048     & \textbf{35/6048}     & 3.9\%       & \textbf{0.6\%}     \\ \hline
\multicolumn{2}{c||}{scans with wrong bifurcation}                                                       & 28/400       & \textbf{11/400}      & 7.0\%       & \textbf{2.8\%}     \\ \hline
\multicolumn{2}{c||}{CL length (mm)}                                                    & 308.9        & \textbf{314.3}       & -           & -         \\ \hhline{==#==|==}
\multicolumn{2}{c||}{overall success rate}                                                    & 217/400      & \textbf{355/400}     & 54.3\%      & \textbf{88.8\%}    \\ \hline
\end{tabular}
\end{table}

Table \ref{tab:comp} shows a detailed analysis regarding the difference between results generated by DeepCL and baseline. Bold items show the method with better performance on each metric. DeepCL finds more branch endpoints on both branch-level and patient-level. The number of wrong bifurcations shown in DeepCL is also less than that in baseline. Besides, DeepCL increased the average patient-level centerline length, due to finding of more endpoints and staying to the center instead of taking straight shortcuts at tortuous regions. Overall, the percentage of scans with successful centerline extraction without any type of error on any branch is substantially increased from 54.3\% to 88.8\%. 

%%%%%%%%%%%%%%%%%%%%%%%%%%%%%%%%%%% Figure %%%%%%%%%%%%%%%%%%%%%%%%%%%%%%%%%%%%%

%%%%%%%%%%%%%%%%%%%%%%%%%%%%%%%%%%%%%%%%%%%%%%%%%%%%%%%%%%%%%%%%%%%%%%%%%%%

%%%%%%%%%%%%%%%%%%%%%%%%%%%%%%%%%%% Figure %%%%%%%%%%%%%%%%%%%%%%%%%%%%%%%%%%%%%
\begin{figure}[t]
\includegraphics[width=\textwidth]{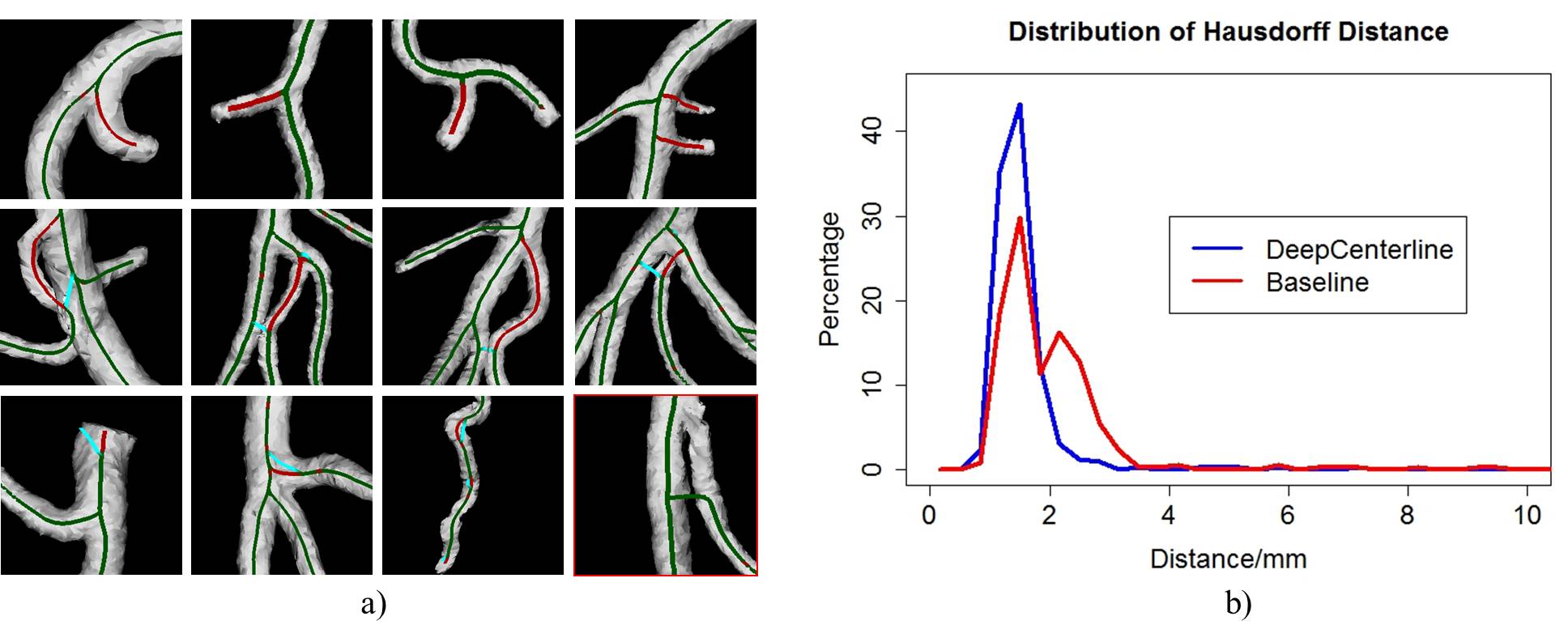}
\caption{Visual comparison of two methods and Hausdorff distance distributions. a) Red is centerlines from DeepCL. Cyan is centerlines from baseline method. Green shows the overlap of both centerlines. First row: DeepCL finds branches missed by baseline method. Second row: DeepCL avoids wrong bifurcations generated by baseline. Third row: DeepCL generates centerlines well-positioned at central location, avoiding taking straight shortcuts at complex bifurcation regions or tortuous segments. The last figure with red border shows a failure case for both DeepCL and baseline. b) Hausdorff distance distribution from voxels to the nearest centerline points for both methods.} \label{fig:visual}
\end{figure}
%%%%%%%%%%%%%%%%%%%%%%%%%%%%%%%%%%%%%%%%%%%%%%%%%%%%%%%%%%%%%%%%%%%%%%%%%%%

Fig. \ref{fig:visual}(a) shows qualitative comparison of both methods. Compared to baseline, DeepCL shows significant improvement in finding more endpoints, reducing number of wrong centerline bifurcations at region with vessels close together, and staying at center instead of taking straight shortcut at regions with high curvature. 

Fig. \ref{fig:visual}(b) shows the distribution of patient-level Hausdorff distance from any segmentation mask voxel to centerline. 
%It indicates how well the centerline is representing the segmentation mask. 
A smaller Hausdorff distance means that \emph{all} voxels in the segmentation mask are ``covered'' by a closer nearby centerline point. The majority of Hausdorff distances for DeepCL centerlines form a peak around 1.7 mm. In contrast, the baseline method has a longer tail towards higher Hausdorff distance values, with a significant percentage of scans having Hausdorff distance above 2 mm. This shows that DeepCL covers all segmentation mask voxels more closely.

%\zzz{I did not have enough time to read all carefully but I think that you may still have the same validation issue I mentioned in teh Abstract and that this issue - solved the way as you propose - is not explaining sufficiently enough. I think that this sis the biggest issue of this manuscript = not the work itself, but the way how you describe the validation and what forms the independent standard. You should try to make it clear.}

\vspace{-10pt}
\subsubsection{Importance of attention}

%%%%%%%%%%%%%%%%%%%%%%%%%%%%%%%%%%% Figure %%%%%%%%%%%%%%%%%%%%%%%%%%%%%%%%%%%%%
\begin{figure}[t]
\includegraphics[width=\textwidth]{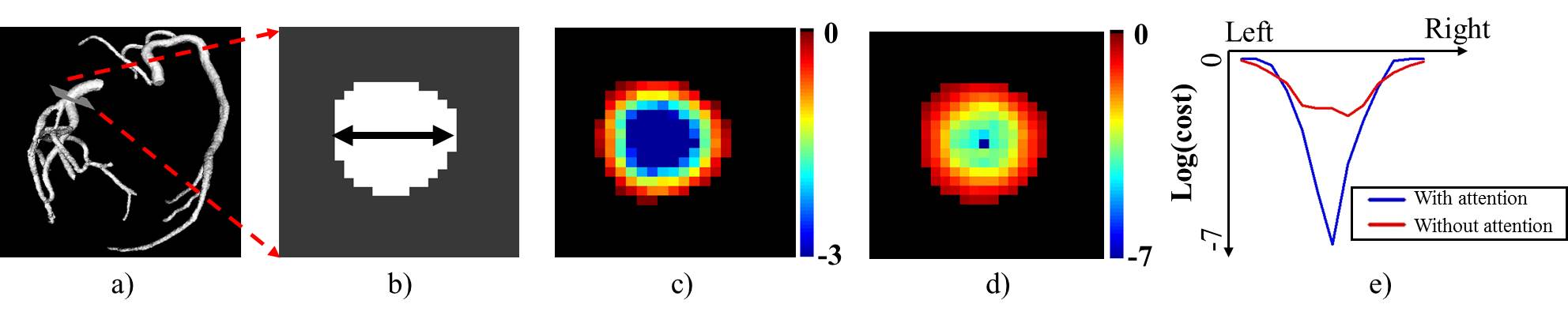}
\caption{Comparison of centerline distance map prediction with and without attention. a) Coronary artery segmetnation mask. b) A cross-sectional view of segmentation mask. c) Centerline distance map without attention module. d) Centerline distance map with attention module. e) Centerline distance map values at the profile line shown as double-arrowed line in b). With attention, the centerline distance map shows a high peak around centerline instead of a plateau by the model without attention. }
%Predicted cost metric for b) without attention module during training. d) Predicted cost metric for b) with attention module during training. e) Cost value distributions for c) and d) at locations shown along bi-directional arrow in b). Red line: no attention module added. Blue line: attention module added as described in DeepCL.} 
\label{fig:att}
\end{figure}
%%%%%%%%%%%%%%%%%%%%%%%%%%%%%%%%%%%%%%%%%%%%%%%%%%%%%%%%%%%%%%%%%%%%%%%%%%%

Fig. \ref{fig:att} compares the centerline distance maps generated with and without the attention module.
%on one cross-sectional view of the segmentation mask. 
As shown in Fig. \ref{fig:att}(e), the centerline distance map has a clear peak around the central location by using attention. In contrast, the CL distance map generated by model without attention results in a ``plateau'' for a large area. 
%From the intensity profile of the artery mask as shown in Fig. \ref{fig:att}(e), shape contrast around the centerline point was obtained with the attention module. On the contrary, a ``plateau'' appeared at a large area around the centerline point without the attention module. 
%When fed into minimal path algorithm, such central centerline distance map plateau causes the resulting centerline to easily pick any point on the plateau.
If this situation occurred at regions with high curvature, then the minimal path extractor can easily pick a straight shortcut passing through non-center plateau points as the resulting centerline. This problem is greatly alleviated by utilizing the attention module to improve the centerline distance map contrast around real centerline point. 

\section{Discussion}
\vspace{-5pt}
The proposed method tackles multiple long-existing challenges of centerline extraction. 
%The proposed model tackles the previously mentioned challenges of centerline extraction in the following way. 
A novel branch endpoint detection algorithm using Gaussian-field based endpoint confidence map is developed to detect the extremely sparse branch endpoints. The centerline distance map is made scale-invariant to the substantial diameter change of vessel branches from proximal to distal sections through local normalization within each cross-sectional view. 
The scale-invariant centerline distance map helps generate well-positioned centerline throughout all sections. Log-transform and attention module are utilized to further highlight the central locations, aiding accurate localization of the single-pixel-wide centerline. The large model capacity of FCN provides robustness to minor imperfections of segmentation masks.

%We would like to emphasize that FCN is a proper and necessary technique to enable the locally normalized centerline distance map. The locally normalized centerline distance map is hard to be achieved by traditional methods. The process of generating the reference map during training involves extracting cross-sectional views at each centerline point. During testing, obtaining the location and direction of each cross-sectional view is very difficult for every centerline point, which is still unknown. Using FCN with a large capacity and ability to model complex functions help us to predict such a complex output. 

Note that the reference ``true'' centerlines used in the training phase are results generated by the baseline method \emph{without} manual correction, due to the difficulty of manual correction of a single-pixel-wide centerline. Despite of this disadvantage, our method achieves better performance than baseline on the test set. This shows the good generalization ability of the proposed model. A further study topic is to use our current method's output as reference standard to train another FCN model. It would be interesting to see whether this second model would further improve upon the first-generation model. 

\section{Conclusion}
We propose a novel centerline extraction framework which combines a multi-task FCN computing a locally normalized centerline distance map and detecting branch endpoints, with a minimal path extractor. The proposed method is the first deep-learning based centerline extraction method that guarantees single-pixel-wide centerline for a complex tree-structured object. Designed to be robust to substantial scale changes at different locations and minor imperfections of segmentation mask, the proposed method generates centerlines with more complete and closer coverage of segmentation masks without false positive branches. 
%Experiment on the challenging coronary artery centerline extraction application shows the proposed method improves the patient-level success rate of centerline extraction from 54.3\% to 88.8\%. 

%DeepCenterline is the first deep-learning based centerline extraction method generadting
%Our paper proposes the first deep-learning based centerline extraction method generating single-pixel-wide centerline for complex tree-structured object. It handles object with multiple branches, significant scale changes at different locations, and minor imperfections of segmentation mask. The resulting centerline achieves more complete and closer coverage of segmentation mask with less errors caused by substantial scale changes or minor errors in segmentation mask. 

%We proposed a novel two-head FCN model with a minimal path extractor for coronary artery centerline extraction. One head of the FCN generates scale-invariant centerline distance map and the other head detects a complete list of branch endpoints. When compared with a state-of-the-art baseline method, our method was insensitive to radius change along vessels, and robust to branch tortuousity and minor imperfections of segmentation mask.
\vspace{-10pt}
\subsubsection{Acknowledgement} The authors would like to thank Xiaoyang Xu and Bin Ouyang for organizing the dataset. This study has received funding by Shenzhen Municipal Government (KQTD2016112809330877). 
\vspace{-10pt}

%\begin{figure}
%\includegraphics[width=\textwidth]{fig1.eps}
%\caption{A figure caption is always placed below the illustration.
%Please note that short captions are centered, while long ones are
%justified by the macro package automatically.} \label{fig1}
%\end{figure}

%
% the environments 'definition', 'lemma', 'proposition', 'corollary',
% 'remark', and 'example' are defined in the LLNCS documentclass as well.
%

\bibliographystyle{splncs04}
\bibliography{mybibliography}

\end{document}